\documentclass[10pt, conference]{IEEEtran}
\usepackage{graphicx}
\usepackage[caption=false]{subfig}
\usepackage{booktabs}

\usepackage{balance}

\newcommand{\email}[1]{#1}

\usepackage{tikz}

\newcommand\copyrighttext{%
  \footnotesize \textcopyright 2020 IEEE. Personal use of this material is permitted.
  Permission from IEEE must be obtained for all other uses, in any current or future
  media, including reprinting/republishing this material for advertising or promotional
  purposes, creating new collective works, for resale or redistribution to servers or
  lists, or reuse of any copyrighted component of this work in other works.
}
\newcommand\copyrightnotice{%
\begin{tikzpicture}[remember picture,overlay]
\node[anchor=south,yshift=10pt] at (current page.south) {\fbox{\parbox{\dimexpr\textwidth-\fboxsep-\fboxrule\relax}{\copyrighttext}}};
\end{tikzpicture}%
}

\begin{document}

\title{%
  Towards automated kernel selection in machine learning systems: A SYCL case
  study
}
\author{%
  \IEEEauthorblockN{John Lawson}
  \IEEEauthorblockA{Codeplay Software Ltd.\\\email{john@codeplay.com}}
}
\maketitle
\copyrightnotice
\begin{abstract}

  Automated tuning of compute kernels is a popular area of research, mainly
  focused on finding optimal kernel parameters for a problem with fixed input
  sizes.  This approach is good for deploying machine learning models, where the
  network topology is constant, but machine learning research often involves
  changing network topologies and hyperparameters. Traditional kernel
  auto-tuning has limited impact in this case; a more general selection of
  kernels is required for libraries to accelerate machine learning research.

  In this paper we present initial results using machine learning to select
  kernels in a case study deploying high performance SYCL kernels in libraries
  that target a range of heterogeneous devices from desktop GPUs to embedded
  accelerators. The techniques investigated apply more generally and could
  similarly be integrated with other heterogeneous programming systems.  By
  combining auto-tuning and machine learning these kernel selection processes
  can be deployed with little developer effort to achieve high performance on
  new hardware.

\end{abstract}

\begin{IEEEkeywords}
  Auto-tuning; SYCL; GPGPU; Machine learning;
\end{IEEEkeywords}

\section{Introduction}

Modern machine learning systems, especially deep neural networks, rely heavily
on a small number of compute intensive routines including convolutions and matrix
multiplication. As such, improving the performance of these routines by tuning
kernel parameters can have a significant impact on the time required to train
large machine learning models.

Automatic kernel selection for GPUs is a well studied area especially since the
rise of GPGPU programming.  Many auto-tuning techniques concentrate on
achieving maximum performance for a problem with given input sizes, with the
expectation that when deployed these systems will carry out computations on
different data of the same sizes.

Modern machine learning research often involves training models and continually
tweaking the model architecture and hyperparameters to obtain the best results.
As the models keep changing, standard auto-tuning techniques are of limited
use, and only come into their own when the final models are deployed. As such
autotuning techniques in machine learning frameworks tend to be dynamic, doing
trial runs the first time an input size is used and choosing the best
for subsequent runs.

Codeplay~\cite{codeplay} has been developing an ecosystem of libraries to
accelerate machine learning using SYCL~\cite{sycl} and aims to provide
close to optimal performance on a range of compute tasks.
A typical SYCL implementation converts kernels into an intermediate
representation that is bundled with the final SYCL library.
Supporting many different kernel instantiations in these libraries adds
complexity and a cost in terms of library size and build times.

In this paper we discuss machine learning techniques to prune the number of
kernel configurations that need to be provided in a library while preserving
performance, and show that simple decision processes can be used to choose the
best of these kernels at runtime. The discussion is based around how these
techniques apply to a matrix multiply case study provided by
SYCL-DNN~\cite{sycldnn}.

\subsection{Related work}

Auto-tuners such as clTune~\cite{cltune} and Kernel Tuner~\cite{kernel_tuner}
have been developed optimize heterogeneous compute kernels. These tuning systems
establish the best kernel parameters for a given set of inputs, but the whole tuning
process has to be run for any new inputs. This can be partially mitigated by
using machine learning~\cite{ml_autotuning,boosted_trees_tuning} to reduce the
size of the parameter space that must be searched.

Auto-tuning in this form has been applied to
various problems including matrix
multiplication~\cite{note_tuning_gemm,nugteren2018clblast},
convolution~\cite{tuning_convs}, FFTs~\cite{tuning_ffts} and
stencils~\cite{autotuning_stencils,autotuning_3d_stencils}.

\section{A matrix multiply case study}

\begin{figure*}
  \centering
  \includegraphics{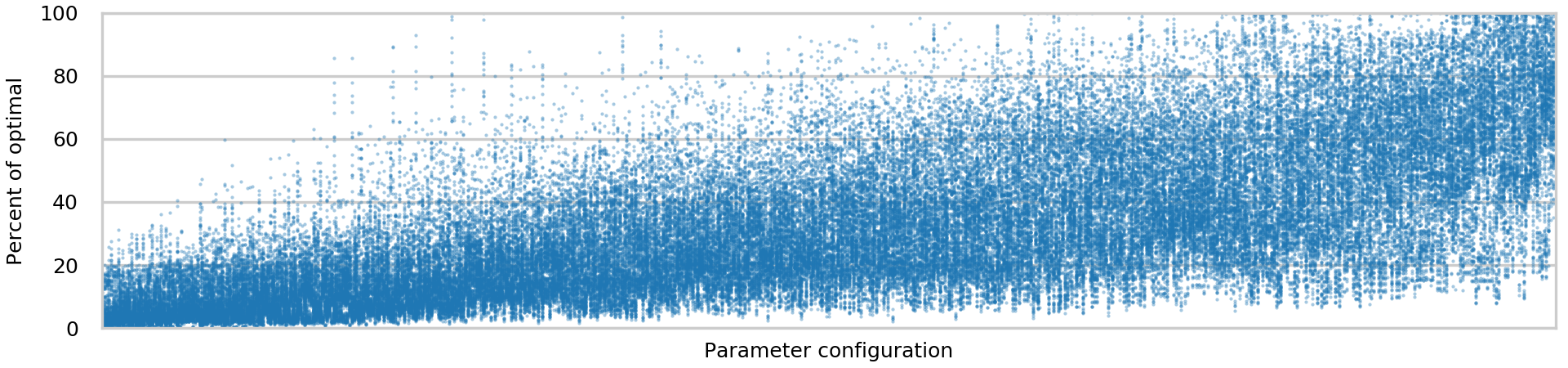}
  \caption{Performance of each configuration for all matrix sizes in the
  dataset, with configurations listed in increasing mean performance. Some
  configurations consistently perform poorly for any set of matrix sizes, while
  those that perform optimally on some sizes still perform poorly on other
  sizes. A few of the configurations that perform poorly on the majority of
  cases can be seen to perform well on a small number of specific matrix sizes.}%
  \label{fig:data_distribution}%
\end{figure*}

A standard kernel tuning example is a general matrix multiply. Such a kernel often
has parameters describing the tile sizes used both at a work group
level for programmatically caching values, and at the work item level with
values in registers. Further parameters include the vector widths used to load
and store values from memory and the sizes of the work groups.

The matrix multiply kernel used in this case study is provided as part of the
SYCL-DNN~\cite{sycldnn} library. SYCL~\cite{sycl} is a
royalty-free, cross-platform parallel programming framework designed to abstract
the complexity of traditional parallel programming models like OpenCL, allowing
developers to use modern C++ features within kernels. In particular, C++
templates are used throughout SYCL-DNN to provide specializations for data
types, tile sizes and other constants with little additional code.

In the matrix multiply kernel each work item computes a tile of the output,
accumulating a given number of values in each step. This gives three compile
time parameters: the two dimensions of the output tile and the size of the
accumulator step. An additional two parameters describe the size of the work
group, however these can be set at runtime and do not require additional kernels
to be compiled.

For each of the tile sizes we considered values of 1, 2, 4 and 8, giving a total
of 64 possible kernels. In addition to this we compared the following work group
sizes: (1, 64), (1, 128), (8, 8), (8, 16), (8, 32), (16, 8), (16,16), (32,8),
(64, 1), (128, 1); giving a total of 640 possible configurations to select from.

Such a small number of configurations allows us to brute-force the performance
for a number of different matrix sizes, however this is not feasible for more
general kernels that have significantly more parameters or for problems where
less strict heuristics are used to limit the number of configurations. For these
cases more complex tuning algorithms have been proposed, such as basin hopping
and evolutionary algorithms; a good discussion of these can be found
in~\cite{kernel_tuner}.

\subsection{The dataset}

Convolutional layers in neural network models can be computed using a matrix
multiply through transformations such as the im2col and Winograd, while fully
connected layers are comprised of a matrix multiply and a bias add.  We
extracted the sizes of matrix multiplies arising from three popular neural
networks: VGG~\cite{vgg}, ResNet~\cite{resnet} and MobileNet~\cite{mobilenetv2},
giving 78, 66 and 26 combinations of matrix sizes to consider respectively (170
combinations total).  For each of these sizes we ran a benchmark for each of the
kernel configurations, recording the runtime of the kernel and number of flops
attained over a number of iterations.  The benchmark platform was an AMD R9 Nano
GPU.

The full dataset is shown in Figure~\ref{fig:data_distribution}, with the
relative performance of each benchmark run for every configuration, sorted by
their overall mean performance. With so many different configurations it is hard
to isolate any individual in the figure, but it is useful to consider the
distribution of performance across the matrix sizes.

There are some configurations that perform badly for all matrix sizes, with
those at the far left never achieving above 30\% of the optimal performance.
The configurations on the far right performed well on average, but still perform
poorly on some matrix sizes. Some configurations in the middle have a low
average performance, but can achieve close to optimal performance on certain
sizes.

\begin{figure}
  \centering
  \includegraphics{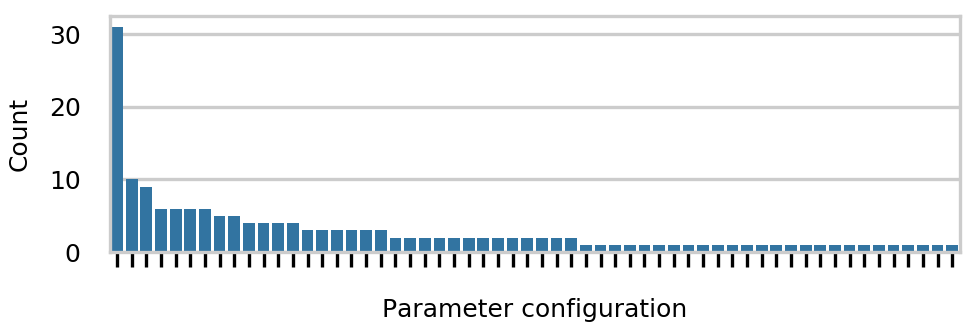}
  \caption{The number of times a configuration of kernel parameters
  achieves optimal performance in
  the dataset. One configuration is best in 32 cases, but 58 distinct
  configurations are best in at least one case.}%
  \label{fig:config_counts}%
\end{figure}

Figure~\ref{fig:config_counts} shows that a single configuration performs best more
than 3 times as often as the next best configuration, however across the full
dataset there are 58 different configurations that give optimal performance for
at least one set of matrix sizes. This long tail causes problems with trying to
prune the number of configurations as it is hard to justify which of the
configurations should be chosen to include in the selection process.

The dataset and the corresponding code is
available online~\cite{jwlawson_kernel_tuner}. The machine learning routines
were provided by scikit-learn~\cite{scikit-learn}.

\subsection{Determining the target number of configurations}

In order to effectively deploy SYCL kernels in a library, only a restricted
number of kernels can be provided without significantly inflating library size.
Principal component analysis (PCA)~\cite{pca_original,prob_pca} can be used to
help determine a good number of kernels to provide to cover the majority of
cases encountered in the dataset.

PCA provides a coordinate system that concentrates the variance of the
dataset in as few components as it can. By comparing the number of components
required to account for a given threshold of the total variance we can estimate
how many different clusters would be required to effectively encompass this
variance in the dataset.

\begin{figure}
  \centering
  \includegraphics{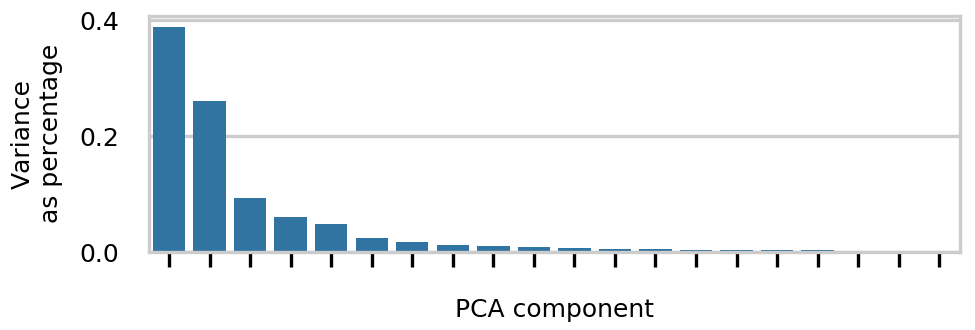}
  \caption{The percentage of the variance of the dataset accounted for by
  each PCA component. Over 80\% of the variance is accounted for in
  the 4 main components, 90\% is accounted for in 8 components, and 95\% in 15.}%
  \label{fig:pca_variance}%
\end{figure}

Figure~\ref{fig:pca_variance} shows the ratio of variance accounted for by each
of the components identified by PCA\@. The first 4 components account for over
80\% of the variance, 8 components account for 90\% and 15 account for 95\%, and
so we investigate limiting the number of kernels between 4 and 15.

\section{Configuration pruning approaches}\label{sec:pruning}

We compared five approaches to pruning the available kernel configurations.  The
simplest pruning method is choosing the top $N$ configurations that obtained
optimal results.

Alongside this we compared methods of clustering the vectors of normalized
performance to give a set of representatives. For each set of matrix sizes (the
input features) in the dataset we have a vector of 640 normalized performance
scores coming from the 640 different kernel configurations. The assumption made
in this paper is that these vectors contain enough structure to provide a good
basis for pruning the number of kernel configurations. Each of the following
techniques uses clustering to establish a set of representatives from the
dataset that exhibit different performance behavior. The kernel configurations
that gives the best performance result for each of the representatives are
chosen as the set of configurations to provide in the compute library.

The first clustering method evaluated is $k$-means clustering, a well known and
relatively simple clustering algorithm that iteratively tries to find clusters
around centroids, and the second is
HDBScan~\cite{hdbscan_paper,hdbscan_software}, a more complex density based
clustering method.  PCA can be used to reduce the dimensionality of the data and
so provide a better coordinate system for $k$-means clustering, which struggles
with high dimensional data. The centroids identified by $k$-means in this new
coordinate system can be mapped back to the original coordinate space to give
representatives of the clusters.

Finally we used a decision tree to do regression on the dataset that maps a set
of matrix sizes to a vector of the expected normalized performance for each
configuration. Limiting the number of leaf nodes in the decision tree ensures
the tree only produces a restricted number of such vectors which are used as the
cluster representatives.

\subsection{The results}

\begin{figure}
  \centering
  \includegraphics{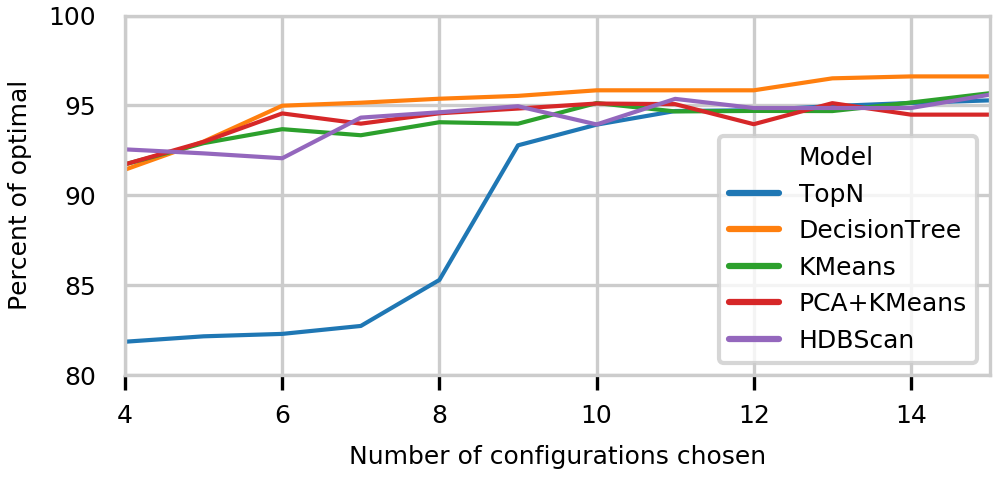}
  \caption{The performance of each pruning technique in
  Section~\ref{sec:pruning} as a percentage of the
  optimal obtainable performance.}%
  \label{fig:error_classes}%
\end{figure}

In order to test how well any of the following approaches generalize to unseen
matrix sizes the data set was randomly segmented into a training and a test
dataset of 136 and 34 elements respectively.

The different kernel configuration selection techniques were trained on the
training dataset to generate a set of kernel configurations limited to a fixed
upper bound. The performance of the clustering technique was measured by taking
the geometric mean of the optimal result achievable given that selection for
each set of matrix sizes in the test set. A score of 100\% would only be
obtained by selecting the best kernel for every matrix size in the test set.

The results are shown in Figure~\ref{fig:error_classes}. When the number of
configurations is very limited, the clustering methods all perform significantly
better than the naive method of selecting the top kernels by count alone. With a
limit of 6 kernels, the decision tree and PCA + $k$-means clustering could both
achieve close to 95\% of the optimal performance. As more configurations were
allowed all techniques improved, achieving around 95\% of the optimal
performance on the test set. The decision tree consistently provided the best
results when 6 or more kernel configurations were allowed, achieving 96.6\% of
the optimal performance in the best case.

\section{Deploying the kernels}\label{sec:classification}

Once a set of kernels has been chosen we still need to provide a process to
select which of these kernels should be run for a set of parameters. Integrating
this process into a library allows the best kernel to be selected that matches
the workload requested by users.

The main challenge in developing a kernel selection process is that it must
balance how good the choice of kernel is with how much time it must spend
in doing the selection. There is little to be gained by choosing a complex
process to achieve slightly better performance if this leads to significantly
more time being spent in that selection process.

Decision trees can be implemented as a series of nested if statements and so are
a good target for deployment provided they do not sacrifice too much
performance. To establish whether decision trees are sufficiently good at
choosing kernel configurations we compare them to other classification methods
that may require more computation to obtain potentially better results.

Table~\ref{tab:classifiers} shows the performance of the various classifiers
tested on the dataset, which include a linear support vector machine (SVM), a
radial SVM, $k$-nearest neighbors and a random forest ensemble.  Despite the
maximum obtainable performance achievable by the kernel configurations chosen
ranging from 93\% to 96.6\% for different numbers of configurations, none of the
models achieve over 89\% performance.

The decision tree outperforms or comes close to the performance of all
other classifiers, except when the model has to choose between
15 configurations.  Overall the classifiers performed worse as the number of
configurations grew, suggesting that the dataset is too small to
correctly learn a generalization encompassing large numbers of classes.

\begin{table}
  \caption{The performance results for the classifiers as a percentage of the
  absolute optimal performance, for the kernel configurations selected by a
  decision tree. Note that the maximum achievable performance for the selection
  of configurations is limited to 92.99\%, 94.98\%, 95.37\% and 96.61\% for the
  5, 6, 8 and 15 configurations respectively.}
  \label{tab:classifiers}
  \centering
  \begin{tabular}{@{}rcccc@{}}
  \toprule
    & \multicolumn{4}{c}{\textbf{Number of configurations}} \\
  \cmidrule(l){2-5}
    \textbf{Classifier} & \textbf{5} & \textbf{6} & \textbf{8} & \textbf{15} \\
  \midrule
DecisionTree & 86.43 & 84.29 & 86.82 & 83.54 \\
RandomForest & 82.99 & 83.70 & 87.99 & 88.13 \\
1NearestNeighbor & 80.45 & 78.44 & 78.30 & 78.21 \\
3NearestNeighbors & 76.41 & 77.95 & 76.34 & 75.45 \\
LinearSVM & 85.88 & 84.17 & 87.96 & 82.50 \\
RadialSVM & 54.95 & 55.01 & 55.01 & 55.01 \\
  \bottomrule
  \end{tabular}
\end{table}

\section{Conclusions}

In this paper we evaluated a number of general methods to prune and deploy
kernel configurations in a compute library, using matrix multiply as a case
study.  Section~\ref{sec:pruning} showed that clustering is effective at
choosing representatives of the kernel configurations that provide good
performance for a range of different input parameters. In particular the use of
a decision tree consistently performed better than the other clustering methods
evaluated.  Section~\ref{sec:classification} showed that a decision tree
choosing the configuration at runtime provides very similar performance to other
selection techniques while being easier to implement in the compute library, so
is a good candidate to deploy in high performance libraries where low latencies
are important.

This work is preliminary with two main areas that will be addressed in the
future.  The
datasets used in this paper are fairly small, causing the models to fail to
generalize which would be mitigated with larger datasets.  The brute-force
techniques used are infeasible for larger problems, where more intelligent
parameter search methods must be used and it is unclear how well the techniques
discussed here generalize to sparse data.

\bibliographystyle{IEEEtran}
\balance
\bibliography{IEEEabrv,bibliography}

\end{document}